%% file: emnlp-ijcnlp-2019.tex
\documentclass[11pt,a4paper]{article}
\usepackage[hyperref]{emnlp-ijcnlp-2019}
\usepackage{times}
\usepackage{latexsym}

\usepackage{url}
\usepackage{amsmath,amssymb}
\usepackage{graphicx}
\usepackage{booktabs}
\usepackage{multirow}
\usepackage[normalem]{ulem}
\graphicspath{{figures/}}

\def\figref#1{Fig.~\ref{#1}}
\def\secref#1{Sec.~\ref{#1}}
\def\tabref#1{Table~\ref{#1}}
\def\eqnref#1{Eqn.~\ref{#1}}
\aclfinalcopy 

\title{Self-Assembling Modular Networks \\ for Interpretable Multi-Hop Reasoning}

\author{Yichen Jiang \and Mohit Bansal \\
  UNC Chapel Hill \\
  {\tt \{yichenj, mbansal\}@cs.unc.edu} \\
 }

\date{}
\begin{document}
\maketitle
\input{tex/abstract}
\input{tex/intro}
\input{tex/related}
\input{tex/models}
\input{tex/experiments}
\input{tex/results}
\input{tex/analysis}
\input{tex/conclusion}

\section*{Acknowledgments}
\vspace{-5pt}
We thank the reviewers for their helpful comments. This work was supported by ARO-YIP Award \#W911NF-18-1-0336, DARPA \#YFA17-D17AP00022 and awards from Google, Facebook, Salesforce, and Adobe (plus Amazon and Google GPU cloud credits). The views are those of the authors and not of the funding agency.

\bibliography{emnlp-ijcnlp-2019}
\bibliographystyle{acl_natbib}

\end{document}

%% file: tex/abstract.tex
\begin{abstract}

Multi-hop QA requires a model to connect multiple pieces of evidence scattered in a long context to answer the question.
The recently proposed HotpotQA~\cite{yang2018hotpotqa} dataset is comprised of questions embodying four different multi-hop reasoning paradigms
(two bridge entity setups, checking multiple properties, and comparing two entities), making it challenging for a single neural network to handle all four.
In this work, we present an interpretable, controller-based Self-Assembling Neural Modular Network~\citep{hu2017learning,hu2018explainable} for multi-hop reasoning, where we design four novel modules (Find, Relocate, Compare, NoOp) to perform unique types of language reasoning. 
Based on a question, our layout controller RNN dynamically infers a series of reasoning modules to construct the entire network.
Empirically, we show that our dynamic, multi-hop modular network achieves significant improvements over the static, single-hop baseline (on both regular and adversarial evaluation).
We further demonstrate the interpretability of our model via three analyses. 
First, the controller can softly decompose the multi-hop question into multiple single-hop sub-questions to promote compositional reasoning behavior of the main network.
Second, the controller can predict layouts that conform to the layouts designed by human experts. 
Finally, the intermediate module can infer the entity that connects two distantly-located supporting facts by addressing the sub-question from the controller.\footnote{Our code is publicly available at:  \url{https://github.com/jiangycTarheel/NMN-MultiHopQA}}
\end{abstract}

%% file: tex/intro.tex
\section{Introduction}
\begin{figure}[t]
\centering
\includegraphics[width=0.47\textwidth]{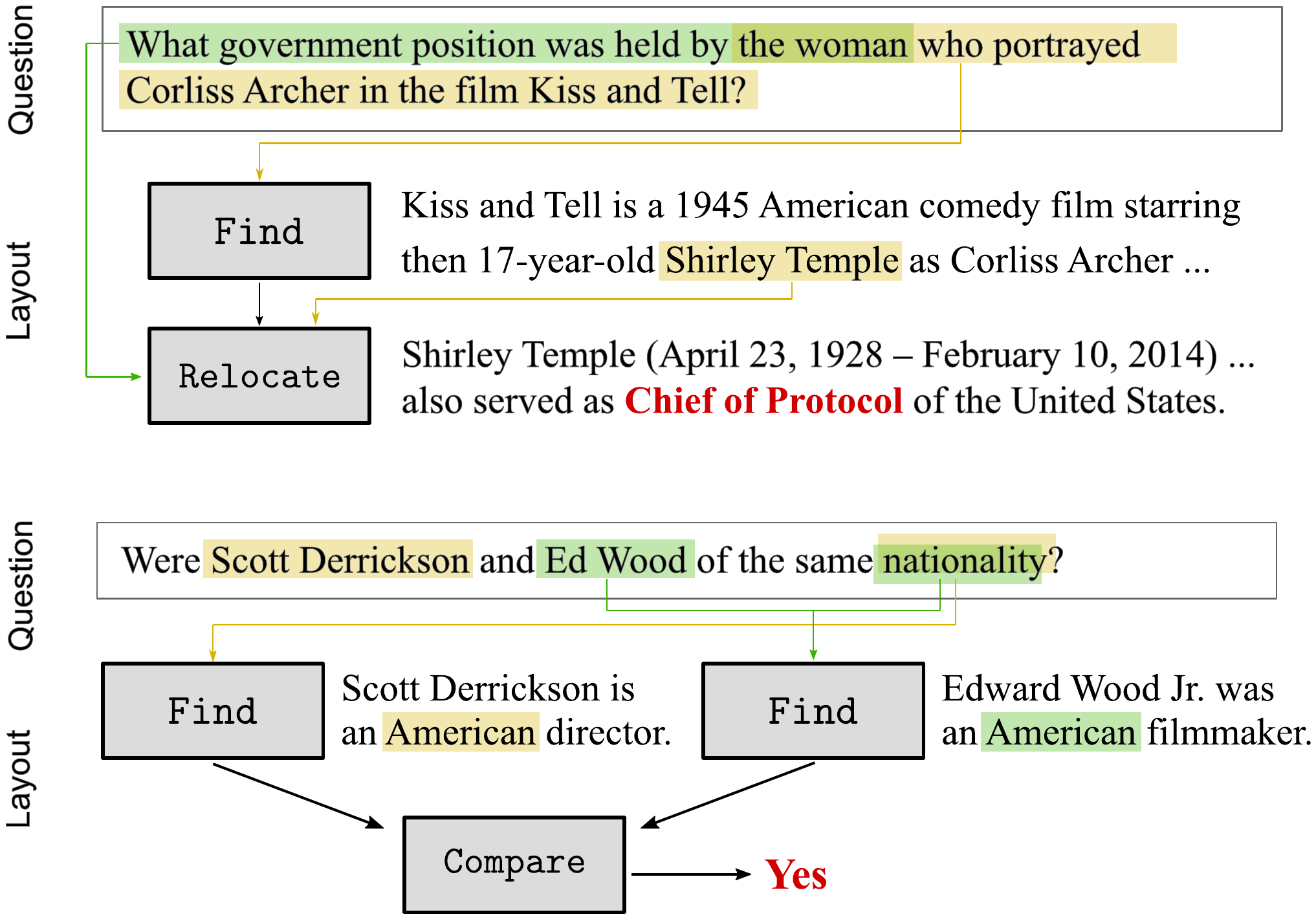}
\vspace{-6pt} 
\caption{Two HotpotQA examples and the modular network layout predicted by the controller.
\vspace{-10pt}
\label{fig:intro_examples}}
\end{figure}

The task of multi-hop question answering (QA) requires the model to answer a natural language question by finding multiple relevant information pieces scattered in a given natural language context.
It has attracted more attention recently and multiple datasets have been proposed, including the recent HotpotQA~\cite{yang2018hotpotqa} that is comprised of questions embodying four different multi-hop reasoning paradigms: inferring the bridge entity to complete the 2nd-hop question (first question in \figref{fig:intro_examples}), inferring the answer through a bridge entity, checking multiple properties to select the answer, and comparing two entities (second question in \figref{fig:intro_examples}).
For the first question, it is necessary to first find the person ``who portrayed Corliss Archer in the film Kiss and Tell", and then find out the ``government position" she held.
For the second question, one may arrive at the answer by finding the country where Scott and Ed are from, and then comparing the two nationalities to conclude whether they are the same.

Multi-hop QA is more challenging than single-hop QA for two main reasons.
First, the techniques used for single-hop QA are not sufficient to answer a multi-hop question.
In single-hop QA tasks like SQuAD~\citep{rajpurkar2016squad}, the evidence necessary to answer the question is concentrated in a short context (Q: ``What is the color of the grass'', Context: ``The grass is green.'', Answer: ``Green'').
Such datasets emphasize the role of matching the information between the question and the short context surrounding the answer (``sky$\rightarrow$sky, color$\rightarrow$blue''), and can be solved by building a question-aware context representation~\citep{seo2016bidaf,xiong2016dynamic}.
In contrast, for multi-hop QA, directly matching the semantics of the question and context leads to the entity that bridges the two supporting facts (e.g., ``Shirley Temple"), or the entities that need to be compared against each other (e.g., nationalities of Scott and Ed).
In both cases, further action is required on top of direct semantic matching in order to get to the final answer.
Second, HotpotQA is comprised of questions of four different types of multi-hop reasoning paradigms.
In \figref{fig:intro_examples}, the first question embodies a sequential reasoning path where the model has to solve a sub-question to get an entity ``Shirley Temple", which then leads to the answer to the main question about her ``government position".
The second question, on the other hand, requires a tree-structured reasoning path where the model first locates two entities on the leaves and then compares them to get the answer.
The difference in the required reasoning skills makes it hard for a single static model to handle all types of questions.

To automatically discover the multiple elaborated reasoning steps as required in HotpotQA, a model needs to dynamically adopt a sequence of different reasoning behaviors based on specific questions, which is still unexplored in the field of large-scale text-based QA.
In our work, we present a highly-interpretable self-assembling Neural Modular Network~\citep{andreas2016learning,hu2017learning} with three novel modules designed for multi-hop NLP tasks: \texttt{Find}, \texttt{Relocate}, \texttt{Compare}, where each module embodies a unique type of reasoning behavior. 
The \texttt{Find} module is similar to the previously-introduced 1-hop bi-attention model~\citep{seo2016bidaf,xiong2016dynamic}, which produces an attention map over the context words given the context and the question representation.
For the first example in \figref{fig:intro_examples}, a \texttt{Find} module is used to find the answer (``Shirley Temple") to the sub-question (``who portrayed Corliss Archer ...").
The \texttt{Relocate} module takes the intermediate answer to the previous sub-question (``Shirley Temple"), combines it with the current sub-question (``What government position was held by ..."), and then outputs a new attention map conditioned on the previous one.
The \texttt{Compare} module intuitively ``compares" the outputs from two previous modules based on the question.
For the second example in \figref{fig:intro_examples}, after previous modules find the nationality of ``Scott" and ``Ed" (both American), the \texttt{Compare} module outputs the answer ``Yes" based on the word ``same" in the question.
We also use the \texttt{NoOp} module that keeps the current state of the network unchanged when the model decides no more action is needed.
After all the reasoning steps are performed, the final attention map is used to predict the answer span, and the vector output from the \texttt{Compare} module is used to predict whether the answer should be a context span or ``Yes/No". 

Next, to dynamically assemble these modules into a network based on the specific question, we use a controller RNN that, at each step, infers the required reasoning behavior from the question, outputs the sub-question, and predicts a soft combination of all available modules. As shown in \figref{fig:intro_examples}, a modular network $\texttt{Relocate(\texttt{Find()})}$ is constructed for the bridge-type question and another network with the layout $\texttt{Compare(\texttt{Find}(), \texttt{Find}())}$ is built for the comparison-type question.
In order to make the entire model end-to-end differentiable in gradient descent, we follow \citet{hu2018explainable} to use continuous and soft layout prediction and maintain a differentiable stack-based data structure to store the predicted modules' output.
This approach to optimize the modular network is shown to be superior to using a Reinforcement Learning approach which makes hard module decisions.
\figref{fig:snmn_model} visualizes this controller that predicts the modular layout and the network assembled with the selected modules.
We further apply intermediate supervision to the first produced attention map to encourage finding the bridge entity (``Shirley Temple" in \figref{fig:intro_examples}). 
The entire model is trained end-to-end using cross entropy loss for all supervision.

Overall, our self-assembling controller-based Neural Modular Network achieves statistically significant improvements over both the single-hop bi-attention baseline and the original modular network~\cite{hu2018explainable} designed for visual-domain QA.
We also present adversarial evaluation~\cite{Jiang2019reasoningshortcut}, where single-hop reasoning shortcuts are eliminated and compositional reasoning is enforced; and here our NMN again outperforms the BiDAF baseline significantly in the EM score (as well as after adversarial training).
We further demonstrate the interpretability of our modular network with three analyses. 
First, the controller understands the multi-hop semantics of the question and can provide accurate step-by-step sub-questions to the module to lead the main network to follow the reasoning path.
Second, the controller can successfully predict layouts that conform to the layouts designed by human experts. 
Finally, before arriving at the final answer, the intermediate module can infer the bridge entity that connects two distantly-located supporting facts by leveraging the step-specific sub-question from the controller.
All of these analyses show that our Modular Network is not operating as a black box, but demonstrates highly interpretable compositional reasoning behavior, which is beneficial in transparent model development and safe, trustworthy real-world applications.
Moreover, our recent experiments show that our NMN with fine-tuned BERT embedding inputs allows for stronger interpretability/modularity than non-modular BERT-style models (and non-BERT NMNs) while also maintaining BERT-style numbers, e.g., it predicts more network layouts that conform to human-expert design than the NMN with GLoVe~\cite{pennington2014glove} embeddings.

In summary, the contribution of this work is three-fold: 1) This is the first work to apply self-assembling modular networks to text-based QA; 2) We design three novel modules to handle multi-hop questions in HotpotQA; 3) The resulting network is interpretable in terms of the model's intermediate outputs and the assembled layout. 

%% file: tex/related.tex
\section{Related Works}
\paragraph{Multi-hop Reading Comprehension}
The last few years have witnessed significant progress on large-scale QA datasets including cloze-style tasks~\citep{Hermann:cnndm}, open-domain QA~\citep{yang2015wikiqa}, and more~\citep{rajpurkar2016squad,rajpurkar2018squad2}.
However, all of the above datasets are confined to a single-document context per question domain. 
\citet{joshi2017triviaqa} introduced a multi-document QA dataset with some questions requiring cross-sentence inferences to answer.
The bAbI dataset~\cite{weston2015babi} requires the model to combine multiple pieces of evidence in the synthetic text. 
\citet{welbl2017qangaroo} uses Wikipedia articles as the context and a subject-relation pair as the query, and constructs the multi-hop QAngaroo dataset by traversing a directed bipartite graph so that the evidence required to answer a query could be spread across multiple documents. 
HotpotQA~\cite{yang2018hotpotqa} is a more recent multi-hop QA dataset that has crowd-sourced questions with more diverse syntactic and semantic features compared to QAngaroo. 
It includes four types of questions, each requiring a different reasoning paradigm. 
Some examples require inferring the bridge entity from the question (Type I in \citet{yang2018hotpotqa}), while others demand fact-checking or comparing subjects' properties from two different documents (Type II and comparison question).
Concurrently to our work, \citet{min2019multi} also tackle HotpotQA by decomposing its multi-hop questions into single-hop sub-questions to achieve better performance and interpretability. However, their system approaches the question decomposition by having a decomposer model trained via human labels, while our controller accomplishes this task automatically with the soft attention over the question-words' representation and is only distantly supervised by the answer and bridge-entity supervision, with no extra human labels.
Moreover, they propose a pipeline system with the decomposers, an answer-prediction model, and a decomposition scorer trained separately on the previous stage's output. Our modular network, on the other hand, is an end-to-end system that is optimized jointly.

\paragraph{Neural Modular Network}
Neural Modular Network (NMN) is a class of models that are composed of a number of sub-modules, where each sub-module is capable of performing a specific subtask.
In NMN~\citep{andreas2016neural}, N2NMN~\citep{hu2017learning}, PG+EE~\citep{johnson2017inferring}, GroundNet~\citep{cirik2018using} and TbD~\citep{mascharka2018transparency}, the entire reasoning procedure starts by analyzing the question and decomposing the reasoning procedure into a sequence of sub-tasks, each with a corresponding module. 
This is done by either a parser~\citep{andreas2016neural,cirik2018using} or a layout policy~\citep{hu2017learning,johnson2017inferring,mascharka2018transparency} that turns the question into a module layout.
Then the module layout is executed with a neural module network. 
Overall, given an input question, the layout policy learns what sub-tasks to perform, and the neural modules learn how to perform each individual sub-task.

However, since the the modular layout is sampled from the controller, the controller itself is not end-to-end differentiable and has to be optimized using Reinforcement Learning Algorithms like Reinforce~\citep{williams1992simple}.
\citet{hu2018explainable} used soft program execution where the output of each step is the average of outputs from all modules weighted by the module distribution, and showed its superiority over hard-layout NMNs.
All previous works in NMN, including~\citet{hu2018explainable} targeted visual question answering, referring expressions, or structured knowledge-based GeoQA, and hence the modules are designed to process image or KB inputs.
We are the first to apply modular networks to unstructured, text-based QA, where we redesigned the modules for language-based reasoning by using bi-attention~\citep{seo2016bidaf,xiong2016dynamic} to replace convolution and multiplication of the question vector with the image feature.
Our model also has access to the full-sized bi-attention vector before it is projected down to the 1-d distribution.

\begin{figure}[t]
\centering
\includegraphics[width=0.47\textwidth]{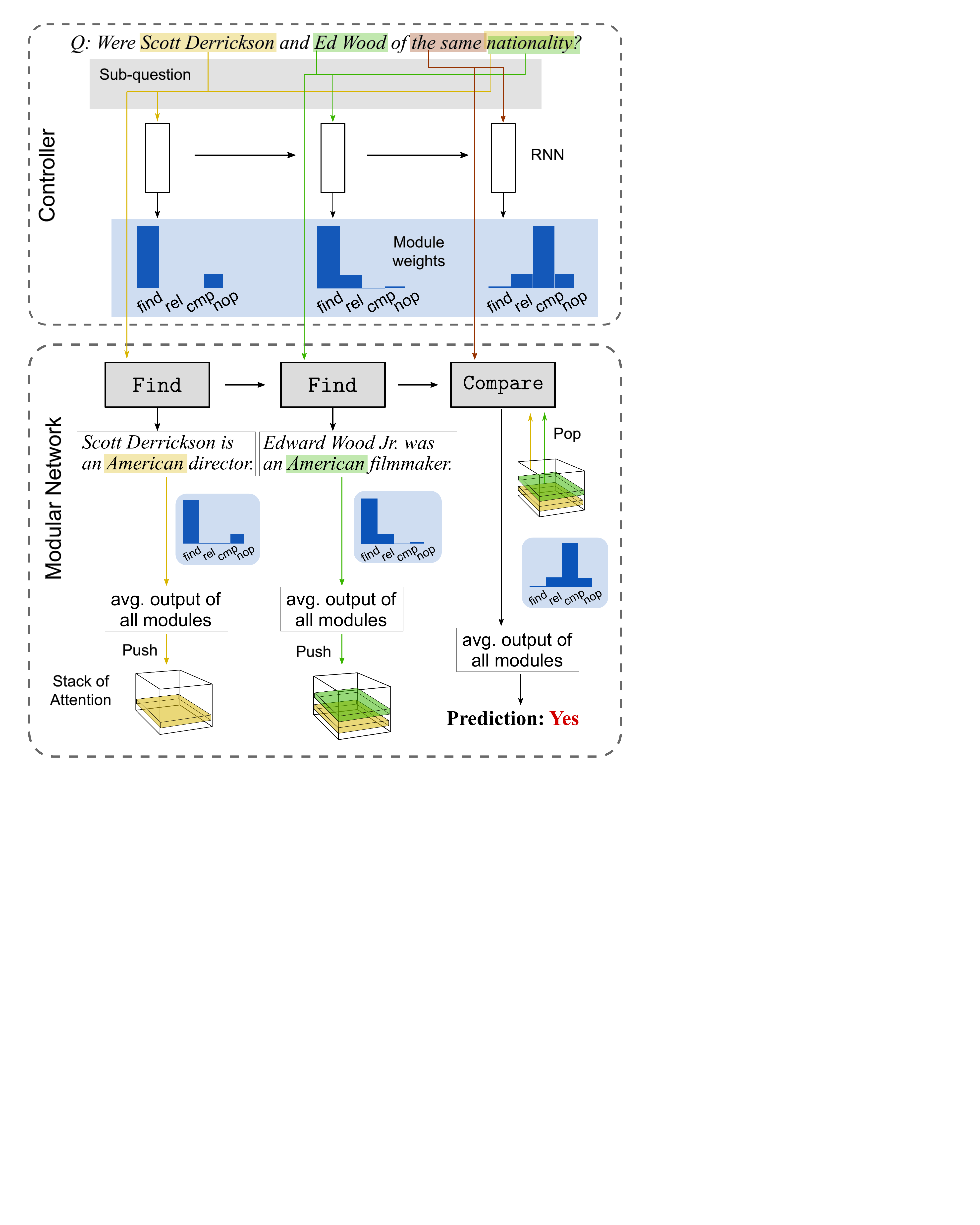}
\vspace{-5pt} 
\caption{Modular network with a controller (top) and the dynamically-assembled modular network (bottom). 
At every step, the controller produces a sub-question vector and predicts a distribution to weigh the averages of the modules' outputs. 
\label{fig:snmn_model} 
\vspace{-15pt}}
\end{figure}

\paragraph{Architecture Learning}
Our work also shares the spirit of recent research on Neural Architecture Search (NAS)~\citep{zoph2016neural,pham2018efficient,liu2018darts}, since the architecture of the model is learned dynamically based on controllers instead of being manually-designed.
However, Neural Architecture Search aims to learn the structure of the individual CNN/RNN cell with fixed inter-connections between the cells, while Modular Networks have preset individual modules but learns the way to assemble them into a larger network.
Moreover, Modular Networks' architectures are predicted dynamically on each data point, while previous NAS methods learn a single cell structure independent of the example.

%% file: tex/models.tex
\section{Model}
In this section, we describe how we apply the Neural Modular Network to the multi-hop HotpotQA task~\citep{yang2018hotpotqa}.
In \secref{ssec:controller} and \secref{ssec:stacknmn}, we describe the controller, which sketches the model layout for a specific example, and introduce how it assembles the network with the predicted modules.
In \secref{ssec:modules}, we go into details of every novel module that we design for this task.

\subsection{Encoding}
We use a Highway Network~\citep{srivastava2015highway} to merge the character embeddings and GloVe word embeddings~\citep{pennington2014glove}, building word representations for the context and the question as \(\mathbf{x} \in \mathbb{R}^{S \times v}\) and \(\mathbf{q} \in \mathbb{R}^{J \times v}\) respectively, where \(S\) and \(J\) are the lengths of the context and the question.
We then apply a bi-directional LSTM-RNN~\citep{hochreiter1997lstm} of $d$ hidden units to get the contextualized word representations for the context and question: $\mathbf{h} = \mathrm{BiLSTM} (\mathbf{x}); \; \mathbf{u} = \mathrm{BiLSTM} (\mathbf{q})$ so that \(\mathbf{h} \in \mathbb{R}^{S \times 2d}\) and \(\mathbf{u} \in \mathbb{R}^{J \times 2d}\). 
We also use a self attention layer~\citep{zhong2019coarse} to get $qv$, a fixed-sized vector representation of the question.

\subsection{Model Layout Controller}
\label{ssec:controller}
In a modular network, the controller reads the question and predicts a series of our novel modules that could be executed in order to answer the given question.
For multi-hop QA, each module represents a specific 1-hop reasoning behavior and the controller needs to deduce a chain of reasoning behavior that uncovers the evidence necessary to infer the answer.
Similar to the model in~\citet{hu2018explainable}, at the step $t$, the probability of selecting module $i$ is calculated as:
\begin{equation} \label{eq:controller_module}
\begin{split}
q_t &= W_{1,t}\cdot qv + b_{1,t}\\
cq_t &= W_{2}\cdot[c_{t-1}; q_t] + b_{2}\\
s_t &= W_{3}\cdot cq_t \\
p_{t,i} &= \mathrm{Softmax} (s_t) \\
\end{split}
\end{equation}
where $c_{t-1}$ is the controller's hidden state and $q_t$ is the vector representation of the question coming out of the encoding LSTM. $W_3$ projects the output to the dimension of $1\times I$, where $I$ is the number of available modules. 
When predicting the next module, the controller also supplies the current module, which we design for the language domain, with a fixed-sized vector as the sub-question at the current reasoning step (the gray-shaded area of the controller in \figref{fig:snmn_model}).
Consider the first example in \figref{fig:intro_examples}. At the first reasoning step, the \texttt{Find} module's sub-question is ``the woman who portrayed Corliss Archer in the film Kiss and Tell".
To generate this sub-question vector, the controller first calculates a distribution over all question words and then computes the weighted average of all the word representations:
\begin{equation} \label{eq:controller_subq}
\begin{split}
e_{t,j} &= W_{4}\cdot(cq_t \cdot u_j) + b_{4}\\
cv_t &= \mathrm{Softmax} (e_t)\\
c_t &= \sum_{j} cv_{t,j} \cdot u_j
\end{split}
\end{equation}
The sub-question vector $c_t$ also serves as the controller's hidden state and is used to calculate the module probability in the next reasoning step.
Similarly, in the second reasoning step, the sub-question of the \texttt{Relocate} module is ``What government position was held by" and the answer from the \texttt{Find} module at the first step.

\subsection{Stack NMN with Soft Program Execution}
\label{ssec:stacknmn}
In our NMN, some modules may interact with the attention output from previous modules to collect information that could benefit the current reasoning step.
For tree-structured layouts, such as $\texttt{Compare}(\texttt{Find}(), \texttt{Find}())$ as shown in \figref{fig:intro_examples}, the latter \texttt{Compare} module requires access to previous outputs. 
We follow \citet{hu2018explainable} to use a stack to store attention outputs generated by our modules.
In the tree-structured layout, the two \texttt{Find} modules push two attention outputs onto the stack, and the \texttt{Compare} module pops these two attention outputs to compare the content they each focus on.
To make the entire model differentiable, we also adopt soft program execution from \citet{hu2018explainable}, where the output of each step is the sum of outputs from all modules weighted by the module distribution (blue-shaded area in \figref{fig:snmn_model}) computed by the controller.

\subsection{NMN Modules}
We next describe all the modules we designed for our Neural Modular Network.
All modules take the question representation $\mathbf{u}$, context representation $\mathbf{h}$, and sub-question vector $\mathbf{c_t}$ as input.
The core modules that produce the attention map over the context are based on bi-attention~\citep{seo2016bidaf} between the question and context, instead of relying on convolution as in previous NMNs~\citep{hu2017learning,hu2018explainable}.
Every module outputs a fixed-size vector/zero vector as the memory state of the NMN, and is also able to push/pop attention maps onto/from the stack.
These modules, each solving a single-hop sub-question, are chained together according to the layout predicted by the controller to infer the final answer to the multi-hop question.

\label{ssec:modules}
\paragraph{\texttt{Find}}
This module performs the bidirectional attention~\cite{seo2016bidaf} between question $\mathbf{u}$ and context $\mathbf{h'} = \mathbf{h}\cdot\mathbf{c_t}$. By multiplying the original context representation by $\mathbf{c_t}$, we inject the sub-question into the following computation.
The model first computes a similarity matrix $M^{J\times S}$ between every question and context word and use it to derive context-to-query attention:
\vspace{-5pt}
\begin{equation} \label{eq:c-to-q}
\begin{split}
M_{j,s} &= W_1\mathbf{u}_j + W_2\mathbf{h'}_s + W_3(\mathbf{u}_j \odot \mathbf{h'}_s)\\
p_{j,s} &= \frac{\mathrm{exp}(M_{j,s})}{\sum_{j=1}^{J}\mathrm{exp}(M_{j,s})}\\
\mathbf{c_q}_s &= \sum_{j=1}^{J} p_{j,s}\mathbf{u}_j
\end{split}
\end{equation}
where $W_1, W_2$ and $W_3$ are trainable parameters, and $\odot$ is element-wise multiplication. Then the query-to-context attention vector is derived as:
\vspace{-5pt}
\begin{equation} \label{eq:q-to-c}
\begin{split}
m_s &= \mathrm{max}_{1\leq s\leq S} \;M_{j,s}\\
p_{s} &= \frac{\mathrm{exp}(m_{s})}{\sum_{s=1}^{S}\mathrm{exp}(m_{s})}\\
\mathbf{q_c} &= \sum_{s=1}^{S} p_{s}\mathbf{h}_S
\end{split}
\end{equation}
We then obtain the question-aware context representation as:
$\mathbf{\widetilde{h}} = [\mathbf{h'}; \mathbf{c_q}; \mathbf{h'} \odot \mathbf{c_q}; \mathbf{c_q} \odot \mathbf{q_c}]$, and push this bi-attention result onto the stack. This process is visualized in the first two steps in \figref{fig:snmn_model}.

\paragraph{\texttt{Relocate}}
For the first example in \figref{fig:intro_examples}, the second reasoning step requires finding the ``government position" held by the woman, who is identified in the first step.
We propose the \texttt{Relocate} module to model this reasoning behavior of finding the answer based on the information from the question as well as the previous reasoning step.
The \texttt{Relocate} module first pops an attention map $att_1$ from the stack and computes the bridge entity's representation $\mathbf{b}$ as the weighted average over $\mathbf{h}$ using the popped attention. 
It then computes a bridge-entity-awared representation of context $\mathbf{h_b}$, and then applies a \texttt{Find} module between $\mathbf{h_b}$ and question $\mathbf{u}$:
\begin{equation} \label{eq:relocate}
\begin{split}
\mathbf{b} &= \sum_{s} att_{1,s} \cdot \mathbf{h}_s\\
\mathbf{h_b} &= \mathbf{h} \odot \mathbf{b}\\
\mathbf{\widetilde{h}} &= \texttt{Find} (\mathbf{u}, \mathbf{h_b}, \mathbf{c_t})
\end{split}
\end{equation}
The output $\mathbf{\widetilde{h}}$ is then pushed onto the stack.

\paragraph{\texttt{Compare}}
As shown in the final step in \figref{fig:snmn_model}, the \texttt{Compare} module pops two attention maps $att_1$ and $att_2$ from the stack and computes two weighted averages over $\mathbf{h}$ using the attention maps.
It then merges these two vectors with $c_t$ to update the NMN's memory state $m$.
\begin{equation} \label{eq:compare}
\begin{split}
hs_1 &= \sum_{s} att_{1,s} \cdot \mathbf{h}_s; \;\; hs_2 = \sum_{s} att_{2,s} \cdot \mathbf{h}_s\\
o_{in} &= [c_t; hs_1; hs_2; c_t\cdot (hs_1-hs_2)]\\
m &= W_1\cdot(\mathrm{ReLU} (W_2 \cdot o_{in}))
\end{split}
\end{equation}
The intuition is that the \texttt{Compare} module aggregates the information from two attention maps and compares them according to the sub-question $c_t$.

\paragraph{\texttt{NoOp}}
This module updates neither the stack nor the memory state. 
It can be seen as a skip command when the controller decides no more computation is required for the current example.

\begin{table*}[t]
\centering
\begin{small}
\begin{tabular}[t]{cc|cc}
\toprule
\multicolumn{2}{c|}{\textsc{Original}} & \multicolumn{2}{c}{\textsc{Mutated}}\\
question & answer & question & answer\\
\midrule
Were Pavel Urysohn and Levin known & \multirow{2}*{No} & Were Henry Cavill and Li Na known  & \multirow{2}*{No}\\ 
 for the \textbf{same} type of work? & &  for the \textbf{same} type of work?  & \\ 
\midrule
Were Pavel Urysohn and Levin known& \multirow{2}*{No} & Were Pavel Urysohn and Levin known  & \multirow{2}*{Yes}\\ 
 for the \textbf{same} type of work? & &  for the \textbf{different} type of work?  & \\ 
\midrule
Is Rohan Bopanna \textbf{older} & \multirow{2}*{Yes} & Is Rohan Bopanna \textbf{younger} & \multirow{2}*{No}\\ 
 than Sherwood Stewart? & & than Sherwood Stewart? & \\ 
\midrule
Was Howard Hawks a screenwriter of more & \multirow{2}*{Yes} & Was Arthur Berthelet a screenwriter of more  & \multirow{2}*{No}\\ 
 productions than Arthur Berthelet? & &  productions than Howard Hawks? & \\ 
\bottomrule
\end{tabular}
\vspace{-5pt}
\caption{Examples of mutated comparison-type questions and answers from HotpotQA training set.\vspace{-7pt}
}
\label{table:data_aug}
\end{small}
\end{table*}

\subsection{Bridge Entity Supervision}
\label{ssec:bridge_sup}
Even with the multi-hop architecture to capture the hop-specific distribution over the question, there is no supervision on the controller's output distribution $c$ about which part of the question is related to the current reasoning step, thus preventing the controller from learning the composite reasoning skill.
To address this problem, we supervise the first \texttt{Find} module to predict the bridge entity (``Shirley Temple" in \figref{fig:intro_examples}), which indirectly encourages the controller to look for question information related to this entity (``the woman who portrayed Corliss Archer...") at the first step. 
Since the original dataset does not label the bridge entity, we apply a simple heuristic to detect the bridge entity that is the title of one supporting document while also appearing in the other supporting document.\footnote{There are two supporting documents per example.}

\subsection{Prediction Layer}
After all the reasoning steps have completed, to predict a span from the context as the answer, we pop the top attention map from the stack, apply self-attention~\citep{yang2018hotpotqa}, and project it down to get the span's start index and end index.
To predict yes/no, we take the final memory ($m$ in \eqnref{eq:compare}) and project it down to a 2-way classifier.
We concatenate the question vector $qv$ and memory $m$ and then project down to a 2-way classifier to decide whether to output a span or yes/no.

\subsection{Optimization}
Previous works on modular network~\cite{hu2017learning,hu2018explainable} optimize the controller parameters $\theta$ jointly with the modules parameters $\omega$ on the training set. 
However, we found that our controller converges\footnote{Generating modular layout with probability near 1.} in less than 20 iterations under this training routine. It is also likely that this setup causes the model to overfit to the single-hop reasoning shortcuts in the training set. 
Hence, to prevent such premature convergence and reasoning-shortcut overfitting, we adopt the 2-phase training strategy that is widely used in Neural Architecture Search~\cite{zoph2016neural,pham2018efficient}.
The first phase updates $\omega$ for an entire epoch, followed by the second phase that updates $\theta$ over the entire dev set.
We alternate between the two phases until the entire system converges.

%% file: tex/experiments.tex
\section{Experimental Setup}

\paragraph{Dataset and Metric}
We conduct our training and evaluation on the HotpotQA~\cite{yang2018hotpotqa} dataset's distractor setting, which has 90,447 training examples (72991 bridge-type questions and 17456 comparison-type questions) and 7,405 dev examples (5918 bridge-type questions and 1487 comparison-type questions).
Since there is no public test set available, we split the original dev set into a dev set of 3,702 examples on which we tune our models, and a test set of 3,703 examples that is only used for the final test results.
Each question is paired with 10 documents, among which 2 are supporting documents that contain the evidence necessary to infer the answer, and the other 8 are distracting documents.
We evaluate our models based on the exact-match and F1 score between the prediction and ground-truth answer.

\paragraph{Data Augmentation}
\label{ssec:data_aug}
We augment the comparison-type questions in the training set to promote the model's robustness on this reasoning paradigm. 
We start with getting the part-of-speech tags and NER tag of the question using Corenlp~\citep{manning2014corenlp}.
For each question whose answer is either ``yes" or ``no", we generate a new question by randomly sampling two titles of the same type (based on POS and NER) from the training set to substitute the original entities in the question and corresponding supporting documents (1st row in \tabref{table:data_aug}).
We then employ three strategies to generate 5,342 extra questions by mutating original questions.
First, if the question contains the word ``same" and the answer is yes or no, we substitute ``same" with ``different" and vice versa (2nd row in \tabref{table:data_aug}). 
Second, we detect the comparative and superlative adjectives/adverbs in the original question, transform them into their antonyms using wordnet, and then transform the antonyms back to their comparative form (3rd row in \tabref{table:data_aug}).
Finally, if the question has a comparative adj./adv., we flip the order of the two entities compared (4th row in \tabref{table:data_aug}). 
In all three cases, the answer to the mutated question is also flipped.

\begin{table}[t]
\centering
\begin{small}
\begin{tabular}[t]{lcccc}
\toprule
& \multicolumn{2}{c}{Dev} & \multicolumn{2}{c}{Test}\\
& EM & F1 &  EM & F1\\
\midrule
BiDAF Baseline & 44.68 & 57.19 & 42.7 & 55.81\\
NMN & 31.04 & 40.28 & 30.87 & 39.90\\
Our NMN & \textbf{50.67} & \textbf{63.35} & \textbf{49.58} & \textbf{62.71}\\
+ Data aug. & 50.63 & 63.29 & 49.46 & 62.59\\
\midrule
- Bridge sup. & 46.56 & 58.60 & 45.91 & 57.22\\
- Relocate & 47.81 & 60.22 & 46.75 & 59.23\\
- Compare & 50.29 & 63.30 & 48.87 & 62.52\\
- NoOp & 49.11 & 61.79 & 48.56& 62.10\\
\bottomrule
\end{tabular}
\vspace{-5pt}
\caption{EM and F1 scores on HotpotQA dev set and test set. All models are tuned on dev set.\vspace{-7pt}
}
\label{table:main}
\end{small}
\end{table}

\paragraph{Training Details}
We use 300-d GloVe pre-trained embeddings~\cite{pennington2014glove}.
The model is supervised to predict either the start and end index of a span or ``Yes/No" for specific questions.
The entire model (controller + modular network) is trained end-to-end using the Adam optimizer~\cite{Kingma2014AdamAM} with an initial learning rate of 0.001.

%% file: tex/results.tex
\section{Results}
\subsection{Baseline}
Our baseline is the bi-attention~\cite{seo2016bidaf} + self-attention model as in \cite{clark2017simple,yang2018hotpotqa}, which was shown to be able to achieve strong performance on single-hop QA tasks like SQuAD~\citep{rajpurkar2016squad} and TriviaQA~\cite{joshi2017triviaqa}. 
Our baseline share the preprocessing and encoding layer with the modular network.

\subsection{Primary NMN Results}
We first present our model's performance on the HotpotQA~\citep{yang2018hotpotqa} dev and test set of our split.
As shown in the first three rows of \tabref{table:main}, our modular network achieves significant improvements over both the baseline and the convolution-based NMN~\citep{hu2018explainable} on our test set.
In \tabref{table:bridge-compare}, we further break down the dev-set performance on different question types,\footnote{The 3 types of question other than ``comparison" are all classified as ``bridge-type" questions by ~\citet{yang2018hotpotqa}} and our modular network obtains higher scores in both question types compared to the BiDAF baseline.

\begin{table}[t]
\centering
\begin{small}
\begin{tabular}[t]{lcccc}
\toprule
& \multicolumn{2}{c}{Bridge} & \multicolumn{2}{c}{Compare}\\
& EM & F1 &  EM & F1\\
\midrule
BiDAF Baseline & 43.17 & 57.74 & 45.26 & 51.73\\
NMN &30.15 & 41.49 & 34.27 & 40.07\\
Our NMN & \textbf{49.85} & \textbf{64.49} & 51.24 & 57.20\\
+ Data aug. & 48.97 & 63.85  & \textbf{54.40} & \textbf{60.03}\\
\midrule
- Bridge sup. & 45.59 & 59.45 & 48.82 & 51.73\\
- Relocate & 46.11 & 60.13 & 51.92 & 58.10\\
- Compare & 49.37 & 64.46 & 49.76 & 56.00\\
- NoOp & 48.46 & 63.46 & 49.63 & 55.26 \\

\bottomrule
\end{tabular}
\vspace{-5pt}
\caption{EM and F1 scores on bridge-type and comparison-type questions from HotpotQA dev set. 
 \vspace{-1pt}
}
\label{table:bridge-compare}
\end{small}
\end{table}

\begin{table}[t]
\centering
\begin{small}
\begin{tabular}[t]{lcccc}
\toprule
Train & Reg & Reg & Adv & Adv\\
Eval & Reg & Adv & Reg & Adv \\
\midrule
    BiDAF Baseline & 43.12 & 34.00 & 45.12 & 44.65 \\
    Our NMN & \textbf{50.13} & \textbf{44.70} & \textbf{49.33} & \textbf{49.25} \\
\bottomrule
\end{tabular}
\vspace{-5pt}
\caption{EM scores after training on the regular data or on the adversarial data from~\citet{Jiang2019reasoningshortcut}, and evaluation on the regular dev set or the adv-dev set.
\vspace{-18pt}\label{tab:adversarial}}
\end{small}
\end{table}

\subsection{Ablation Studies}
\paragraph{Data Augmentation:}
We also conduct ablation studies on our modeling choices.
Augmenting the comparison-type questions in the training set boosts the performance on the comparison-type questions in the dev set (comparing row 3 and 4 in \tabref{table:bridge-compare}) without harming the scores on the bridge-type questions too much.

\paragraph{Bridge-entity Supervision:}
Supervising the bridge entity (described in \ref{ssec:bridge_sup}) proves to be beneficial for the modular network to achieve good performance (row 5 in \tabref{table:main} and \tabref{table:bridge-compare}).

\paragraph{Modules:}
As shown in the 6th to 8th row, removing either the \texttt{Compare}, \texttt{Relocate} or \texttt{NoOp} module also causes drops in the metric scores.
Specifically, removing the \texttt{Relocate} module causes significant degrade in bridge-type questions, which solidifies our claim that relocating the attention based on the inferred bridge entity is important for compositional reasoning.
Similarly, removing the \texttt{Compare} module harms the model's performance on comparison-type questions, suggesting the effectiveness of the module in addressing questions that require comparing two entities' properties.
These results demonstrate the contribution of each module toward achieving a self-assembling modular network with the strong overall performance.

\subsection{Comparison with Original NMN Modules}
One primary contribution of this work is that we adapt neural modular networks (NMN)~\citep{andreas2016learning,hu2017learning,hu2018explainable}, which were designed for visual-domain QA, to text-domain QA by rebuilding every reasoning module.
We substitute convolution and multiplication between question vectors and context features with bi-attention as the basic reasoning component in the \texttt{Find} and \texttt{Relocate}.
Moreover, our model maintains a stack of attention outputs before it is projected down to 1-d, thus enabling skip connections when predicting the answer span. As shown in \tabref{table:main} and \tabref{table:bridge-compare}, our adapted modular network outperforms the original NMN significantly.

\subsection{Adversarial Evaluation}
Multiple previous works~\cite{chen2019understanding,min2019compositional} have shown that models performing strongly on HotpotQA are not necessarily capable of compositional reasoning. \citet{Jiang2019reasoningshortcut} proposed to construct adversarial distractor documents to eliminate the reasoning shortcut and necessitate compositional reasoning on HotpotQA dataset. 
To test whether our modular network can perform robust multi-hop reasoning against such adversaries, we evaluate our models on the adversarial dev set.
The second column of \tabref{tab:adversarial} shows that our NMN outperforms the baseline significantly (+10 points in EM score) on the adversarial evaluation, suggesting that our NMN is indeed learning stronger compositional reasoning skills compared to the BiDAF baseline.
We further train both models on the adversarial training set, and the results are shown in the last two columns of \tabref{tab:adversarial}.
We observe that after adversarial training, both the baseline and our NMN obtain significant improvement on the adversarial evaluation, while our NMN maintains a significant advantage over the BiDAF baseline.

%% file: tex/analysis.tex
\section{Analysis}
In this section, we present three analysis methods to show that our multi-hop NMN is highly interpretable and makes decisions similar to human actions.
Combining the analyses below, we can understand how our model reasons to break the multi-step task into multiple single-step sub-tasks that are solvable by specific modules, and harnesses intermediate outputs to infer the answer.

\subsection{Controller's Attention on Questions}
\begin{figure}[t]
\centering
\includegraphics[width=0.45\textwidth]{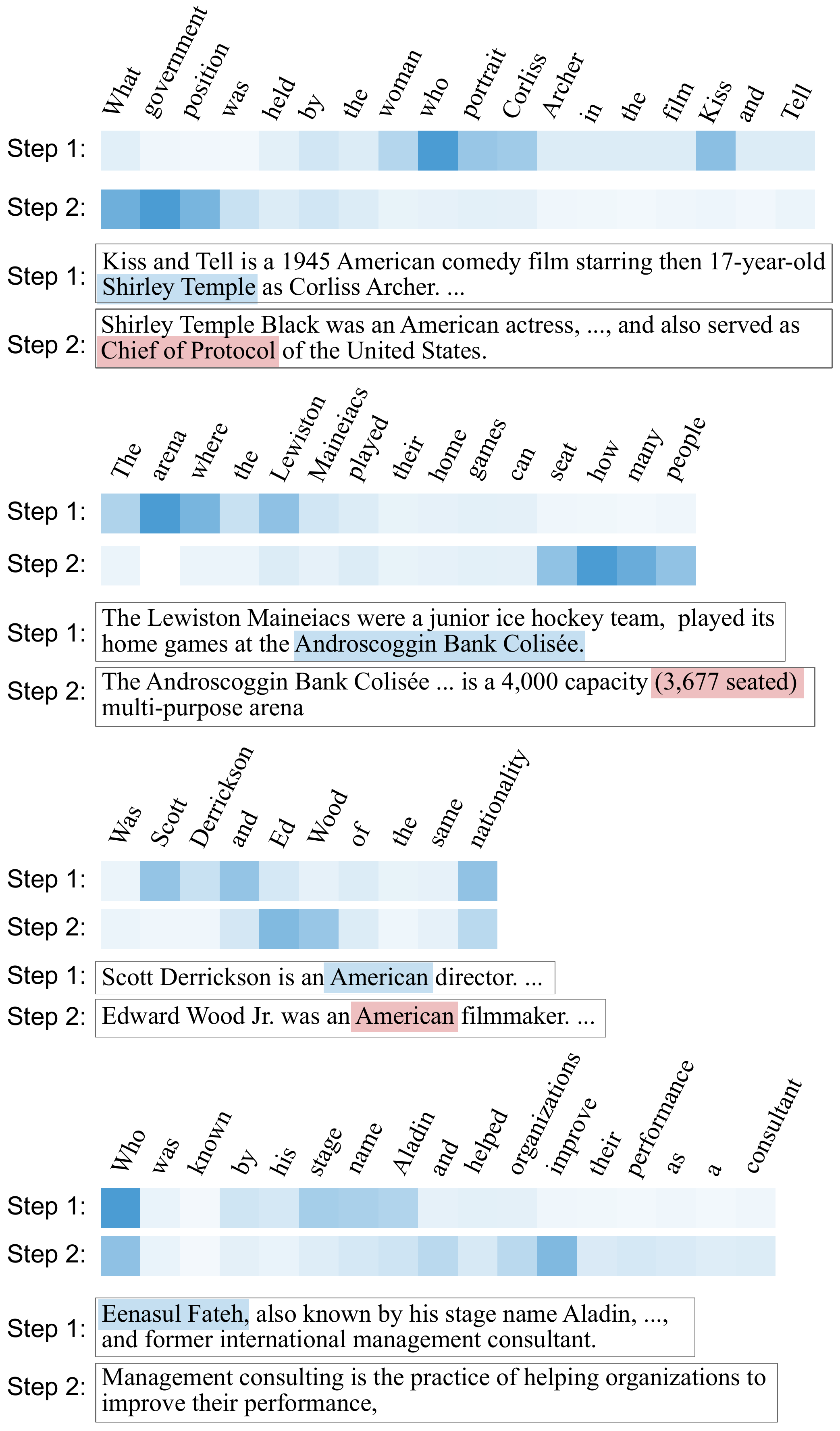}
\vspace{-5pt} 
\caption{The controller's multi-hop attention map on the question (randomly chosen from the first 10 dev examples). 
This attention is used to compute the sub-question representation that is passed to the modules.
\label{fig:controller_attn} 
\vspace{-20pt}}
\end{figure}

Different from single-hop questions, a multi-hop question usually encodes a sequence of sub-questions, among which the final sub-question requires an explicit answer.
Therefore, the first step of solving a multi-hop question is to identify the hidden sub-questions and to sort them according to the correct reasoning order.
Our controller handles this task by computing an attention distribution over all question words (see details in \secref{ssec:controller}) at every reasoning step, which signifies which part of the question forms a sub-question that should be answered at the current step. 
\figref{fig:controller_attn} visualizes four attention maps from the controller when dealing with questions of different reasoning paradigms (randomly chosen from the first 10 dev examples), suggesting that our controller is able to accurately identify the sub-questions in the correct order.
In the first example, the controller is able to attend to the first sub-question ``who portrayed Corliss Archer in ...", and then relocate the attention to the next sub-question ``What government position was held by ..." to complete the 2-step reasoning.

\subsection{Predicting Network Layouts}
\label{ssec:layout}
After the controller decides the sub-question at the current reasoning step, it then predicts a soft layout of the the current step by outputting a distribution $cv$ over all available modules (\texttt{Find}, \texttt{Relocate}, \texttt{Compare}, \texttt{NoOp}). As elaborated in \secref{ssec:stacknmn}, the output vector and attention map of all modules are weighted by $cv$.
To quantitatively evaluate the layout predicted by the controller, we label the expert layout for each question in the dev set based on the dataset's ground-truth question type label  (“\texttt{Find}-\texttt{Relocate}” for bridge-type questions and “\texttt{Find}-\texttt{Find}-\texttt{Compare} / \texttt{Find}-\texttt{Relocate}-\texttt{Compare}” for comparison-type questions).
Next, we convert the soft module predictions into hard modular layouts by picking the module with the largest prediction probability at each step.
Intuitively, we want the controller to assign higher weights to the module that can perform the desired reasoning.
We then compute the percentage of dev set examples where the predicted layout matches the expert layout exactly.
The empirical study shows that the layouts predicted by our NMN controller match the expert layout in upto 99.9\% of bridge-type questions and 68.6\% of comparison-type questions with Yes/No answer (examples shown in \figref{fig:intro_examples} and \figref{fig:snmn_model}).

\subsection{Finding Intermediate Bridge-Entities}
The final action of an intermediate reasoning step is the predicted module reasoning through the question and context to answer the sub-question.
For bridge-type questions (e.g., the first example in \figref{fig:controller_attn}), the required action is to find the name (``Shirley Temple") of ``the woman who portrait Corliss Archer ..." to complete the next sub-question ``What government position was held by Shirley Temple".
For comparison questions (e.g., the third example in \figref{fig:controller_attn}), the model needs to infer the nationalities of ``Scott" and ``Ed" that will be compared in the following step.

\subsection{Recent Results with BERT}
In order to test our Neural Modular Network's interpretability effects in a state-of-the-art setting, we also use the recent pretrained BERT-base~\cite{devlin-etal-2019-bert} model's sequence outputs as stronger input embeddings for our NMN. In order to shorten the length of the context to be processed by BERT, we first fine-tune a BERT-base classification model to select the 2 supporting documents out of 10 documents.
We then retrieve top-3 documents for every example before passing them through the BERT baseline and our BERT+NMN model.
We find that our BERT+NMN model outperforms its corresponding non-BERT counterparts significantly and achieves similar results to the fine-tuned BERT-base model\footnote{Fine-tuned BERT-base model achieves 56.71 EM / 70.66 F1 test scores, and our BERT+NMN obtains 56.63 EM / 71.26 F1 test scores.}, but more importantly this BERT+NMN model allows for stronger interpretability/modularity than non-modular BERT-style models (and non-BERT NMNs), while also maintaining BERT-style numbers. For example, for the human layout evaluation analysis in \secref{ssec:layout}, the controller of our BERT+NMN model can now output ``\texttt{Find}-\texttt{Relocate}" in 99.9\% of bridge-type questions and the stricter requirement of ``\texttt{Find}-\texttt{Find}-\texttt{Compare}" in 96.9\% of comparison-type questions with Yes/No answer, hence predicting substantially more layouts that conform to human-expert design as compared to the non-BERT NMN controller.

%% file: tex/conclusion.tex
\section{Conclusion}
In this work, we proposed a self-assembling neural modular network for multi-hop QA.
We designed three modules that reason between the question and text-based context.
The resulting model outperforms both the single-hop baseline and the original NMN on HotpotQA~\citep{yang2018hotpotqa}.
Because of the interpretable nature of our model, we presented analyses to show that our model does in fact learn to perform compositional reasoning and can dynamically assemble the modular network based on the question. 